\documentclass[twoside,11pt]{article}

\usepackage[hmarginratio=1:1,top=32mm,columnsep=20pt]{geometry} % Document margins

\usepackage{listings}

\usepackage{float}

\usepackage[numbers]{natbib}
\usepackage[caption=false]{subfig}
\usepackage{hyperref}

\usepackage{graphicx}
\DeclareGraphicsExtensions{.png}

\title{libact: Pool-based Active Learning in Python}

\author{Yao-Yuan Yang\thanks{b01902066@ntu.edu.tw},
       Shao-Chuan Lee\thanks{b01902010@csie.ntu.edu.tw},
       Yu-An Chung\thanks{b01902040@csie.ntu.edu.tw},
       Tung-En Wu\thanks{r00942129@ntu.edu.tw},
       Si-An Chen\thanks{r05922089@ntu.edu.tw},\\
       and Hsuan-Tien Lin\thanks{htlin@csie.ntu.edu.tw} \\
       \\
       Department of Computer Science and Information Engineering,\\
       National Taiwan University, Taipei, Taiwan}

\date{}

\begin{document}

\maketitle

\begin{abstract}%   <- trailing '%' for backward compatibility of .sty file
\textsc{libact} is a Python package designed to make active learning easier for
general users. The package not only implements several popular active
learning strategies, but also features the active-learning-by-learning
meta-algorithm that assists the users to automatically select the best strategy
on the fly. Furthermore, the package provides a unified interface for
implementing more strategies, models and application-specific labelers. The
package is open-source on Github, and can be easily
installed from Python Package Index repository.
\end{abstract}

\section{Introduction} \label{sec:intro}

\textsc{libact} is a Python package that provides an easy-to-use
environment for solving active learning problems. In recent years,
several machine learning packages like \textsc{scikit-learn}
\citep{scikit-learn} successfully attract many users by providing
unified interfaces for accessing a wide range of machine learning algorithms.
To the best of our knowledge, there is yet to be a similar package for
active learning in Python. \textsc{libact} follows \textsc{scikit-learn} to
design a framework of unified interfaces for active
learning. The design is non-trivial as active learning problems come with an
interactive and more sophisticated learning protocol than traditional
unsupervised and supervised learning problems.

%% Unlike traditional supervised learning problems that are simply used for training
%% and predicting, active learning problems have a more complicated usage scenario.
%% Especially in the labeling stage, different problems require different labeling processes.
%% For example, labels retrieved from other program require an application interface and
%% problems that depend on human experts need a human computer interface.
%% Without a unified interface, it would be difficult for users to experiment with
%% different algorithms and labeling processes like they would do for supervised learning problems.
%% Therefore, we designed \textsc{libact} as a framework to provide unified interfaces that combine all of the necessary components for active learning.

In \textsc{libact}, we
implement many state-of-the-art active learning algorithms, while
leaving room in the interface for extension to other algorithms. Furthermore,
We address a common user need of algorithm/parameter selection during
active learning by implementing the
active-learning-by-learning (ALBL) meta-algorithm
\citep{hsu2015active}. ALBL can smartly validate several different active
learning algorithms on the fly, and matches the best of those algorithms in
performance. The inclusion of ALBL greatly facilitates the users in terms of
automatic algorithm/parameter selection.

Concrete applications, such as a recent paper on
medical concept recognition~\citep{stanovsky-gruhl-mendes:2017:EACLlong},
have started adopting \textsc{libact}, which is designed
to continuously grow with user needs.
We provide a mature development environment that includes
an issue tracker, continuous integration, unit testing, automatic code analyzer
and document generator. 
New developers can easily join the
\textsc{libact} project
%with code quality controlled by the environment. The environment of \textsc{libact} makes it easier
to spread future research works on active learning for more users.

\section{Interfaces and Usage}

We consider a pool-based active learning problem, which consists of a
set of labeled examples, a set of unlabeled examples, a supervised learning
model, and an labeling oracle \citep{settles2010active}. In each iteration of
active learning, the algorithm (also called a query strategy) queries the oracle
to label an unlabeled example for the model.
The goal is to improve the model rapidly with only a few queries. Based on the components above, we
designed the following four interfaces for \textsc{libact}, which allow
the users to easily try different active learning algorithms, learning models
or labeling oracles for their needs.
%% swap the active learning
%% algorithm, learning model or oracle to the one that fits their problem
%% without modifying other parts of the code.

\paragraph{Dataset.} The \texttt{Dataset} class maintains both the labeled and unlabeled examples.
The examples are stored in an attribute named \texttt{data}, which is a Python list of \texttt{(feature, label)} tuples.
The index of the example in \texttt{data} is used as its identifier.
For an unlabeled example, the \texttt{label} field is assigned with \texttt{None} type.

After retrieving the label for a certain example from the oracle, the \texttt{Dataset.update} method takes in the example index and
its newly retrieved label, and replaces the original entry of
\texttt{(feature, None)} with \texttt{(feature, retrieved\_label)}.
The \texttt{Dataset} class also maintains a list of callback functions that are
triggered each time an example is updated, and the callback functions are mainly used for active learning algorithms that need to update their
internal state after querying the oracle.
Users may also register new callback functions with the \texttt{Dataset.on\_update} method
if there are other needs for their applications.

\paragraph{QueryStrategy.}
The \texttt{QueryStrategy} class is the interface for active learning algorithms. Each \texttt{QueryStrategy} object is associated with a \texttt{Dataset} object. When a \texttt{QueryStrategy} object is initialized, it will automatically register its \texttt{QueryStrategy.update} method as a callback function to the associated \texttt{Dataset} to be informed of any \texttt{Dataset} updates.
The key method that decides the example to be queried is \texttt{QueryStrategy.make\_query}, which returns the identifier of an unlabeled example.
%% When it is called, it has to return an entry ID that represents the example
%% that this active learning algorithm thinks it should query for its label.
By overriding the key method, \textsc{libact} currently implements a diverse spectrum of algorithms as sub-classes of \texttt{QueryStrategy}:

\begin{itemize}
\item Binary Classification:
%\vspace*{-.5em}
%\begin{itemize}
%  \setlength\itemsep{0em}
%  {\small
  Density Weighted Uncertainty Sampling \citep{Nguyen:2004:ALU:1015330.1015349},
  Hinted Sampling with SVM \citep[HintSVM,][]{hintsvm},
  QBC \citep{qbc},
  QUIRE \citep{quire},
  Random Sampling (as a baseline),
  Uncertainty Sampling \citep[][with multi-class support]{lewis1994sequential},
  Variance Reduction \citep{variance2007reduction},
  and ALBL \citep[for algorithm/parameter selection, as discussed in Section~\ref{sec:intro},][]{hsu2015active}
%\vspace*{-.5em}
%\end{itemize}

\item Multi-class Classification:
%\vspace*{-.5em}
%\begin{itemize}
%  \setlength\itemsep{0em}
%  {\small
  %  \item
  Active Learning With Cost Embedding \citep[][with cost-sensitive support]{Huang2016alce},
  Hierarchical Sampling \citep{dasgupta2008hierarchical},
  Expected Error Reduction \citep{settles2010active}
%    }
%\vspace*{-.5em}
%\end{itemize}

\item Multi-label Classification:
%\vspace*{-.5em}
%\begin{itemize}
%  {\small
%  \setlength\itemsep{0em}
  %  \item
  Adaptive Active Learning \citep{li2013active},
  Binary Minimization \citep{brinker2006active},
  Maximal Loss Reduction with Maximal Confidence \citep{yang2009effective},
  Multi-label AL With Auxiliary Learner \citep{CH2011}
%  }
%\end{itemize}
\end{itemize}

\paragraph{Model.} The \texttt{Model} class is the interface for supervised learning models.
%It should contain the implementation of the supervised learning model.
The interface mimics the one in \textsc{scikit-learn} \citep{scikit-learn} to give
easy access to many popular learning models implemented in other Python packages.
The \texttt{Model.train} method takes the labeled examples from a \texttt{Dataset}
and learns from the labeled examples;
the \texttt{Model.predict}, \texttt{Model.predict\_real} and \texttt{Model.predict\_proba} methods output the label predictions, the confidence levels, and the probability estimates on some feature vectors, respectively;
the \texttt{SklearnAdapter} and \texttt{SklearnProbaAdapter} classes convert \textsc{scikit-learn} models for \textsc{libact} to use.

\paragraph{Labeler.} The \texttt{Labeler} class represents the oracle in the
given active learning problem and can be easily customized to support different
applications.
%% For instance, a visual application with human labelers may need to display the given example as a picture to the labelers.
%% For human annotation alone, it requires different ways to show the feature
%% of a given example as a figure, in a graph, or other additional information about
%% the physical meaning of each feature.
Custom-made \texttt{Labeler} should override the \texttt{Labeler.label} method, which
takes in an unlabeled example and returns the retrieved label. As concrete examples, \textsc{libact} include two labelers, \texttt{IdealLabeler}
and \texttt{InteractiveLabler}. \texttt{IdealLabeler} simulates what research papers do when conducting experiments: using a fully-labeled dataset as the backbone of the oracle.
\texttt{InteractiveLabler} serves as a human computer interface that
shows the feature vector of a given example as an image on the screen and fetches the human-provided label through the command line.
%% For inatance, for labeling digits by human, the \texttt{label} method may show the example as a figure on the screen to the human annotator, and the annotator  be given pixels of
%% scanned image from handwritten digits as example.
%% Then the it may show the given example as a figure on the screen and
%% let the human annotator input the corresponding digit through command line.

\paragraph{Usage.}
With the interfaces of \textsc{libact}, the high-level usage pseudo code can be shown below.%
\footnote{See \texttt{examples/plot.py} of \textsc{libact} repository for runnable code.}
The first four lines declare the necessary components in the problem,
and can be replaced with concrete implementations of \texttt{QueryStrategy},
\texttt{Labeler} and \texttt{Model}. Within lines 5 for
the \texttt{quota} iterations of pool-based active learning,
lines 6 to 9 implement the usual query and update steps.
%line 6 uses the \texttt{make\_query} method to get the \texttt{query\_id}
%suggested by \texttt{query\_strategy};
%line 7 sends the example at \texttt{query\_id} to \texttt{labeler} for labeling;
%line 8 updates \texttt{dataset} with newly-labeled example,
%and line 9 improves \texttt{model} with newly-updated \texttt{dataset}.

%% the active learning algorithm for the label. Line 8 extracts the feature of the example
%% from \texttt{dataset} and sends it to the labeler \texttt{labeler}. Lines 9 and 10
%% update the suggested example in \texttt{dataset} with the newly retrieved
%% label \texttt{lbl} and train the learning model \texttt{model}
%% with the updated \texttt{dataset}.

\begin{lstlisting}[language=Python,
                   basicstyle=\small,
                   %                   caption={Example usage of \texttt{libact}},
                   xleftmargin=.2\textwidth,
                   numbers=left]
dataset = Dataset(X, y)
query_strategy = QueryStrategy(dataset)
labeler = Labeler()
model = Model()
for _ in range(quota):
    query_id = query_strategy.make_query()
    lbl = labeler.label(dataset.data[query_id][0])
    dataset.update(query_id, lbl)
    model.train(dataset)
\end{lstlisting}

\begin{figure}[t]
\centering

\subfloat[heart]{\includegraphics[width=.32\textwidth]{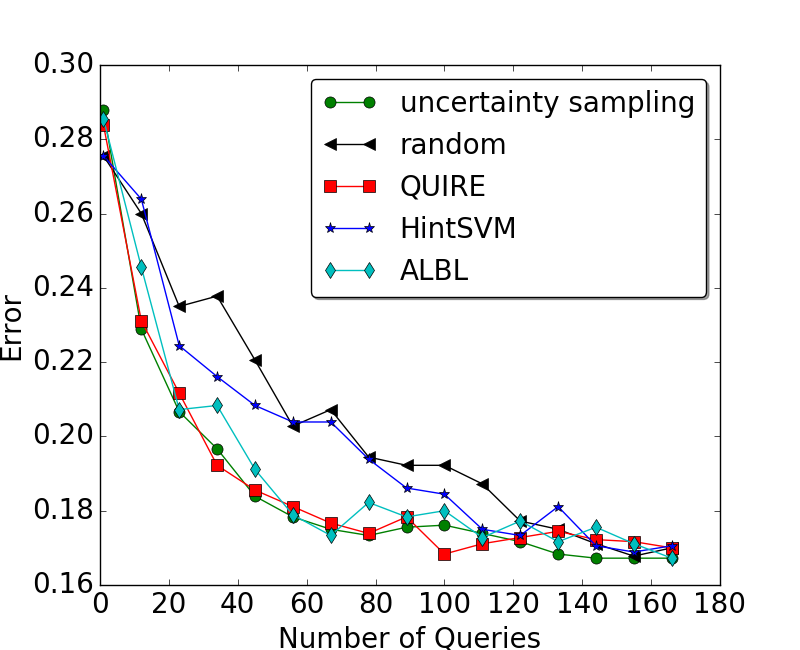}}
\subfloat[australian]{\includegraphics[width=.32\textwidth]{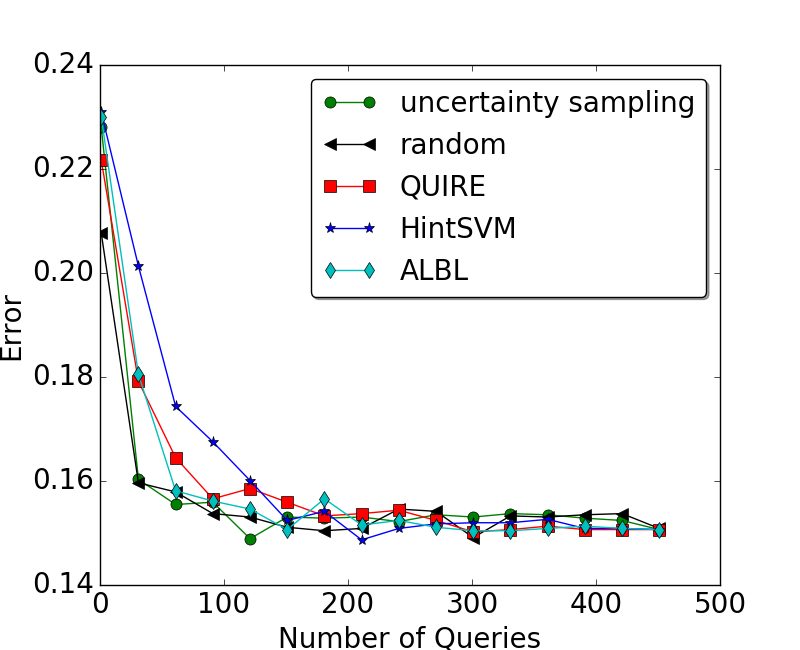}}
\subfloat[diabetes]{\includegraphics[width=.32\textwidth]{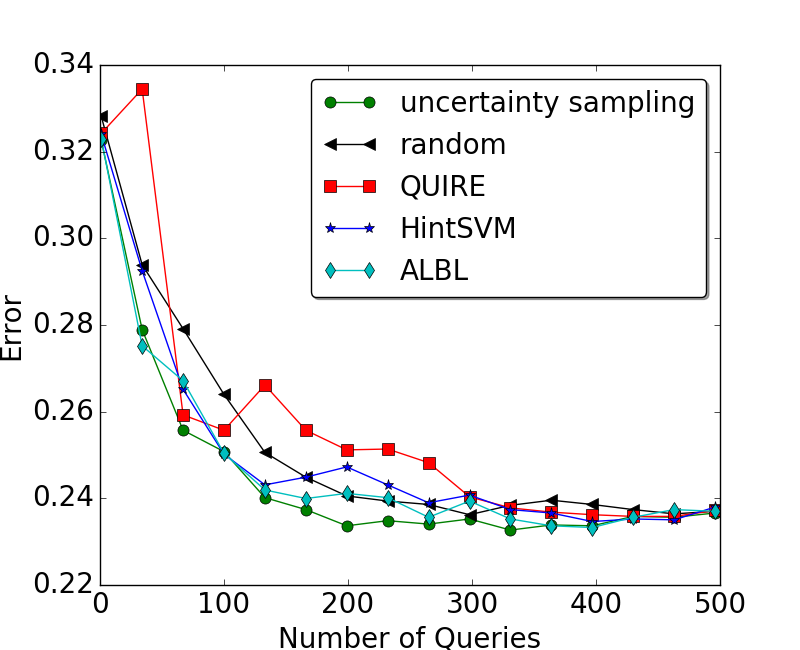}}

\vskip -0.05in
\caption{Comparison between ALBL and other strategies}
\label{fig:albl}
\end{figure}

%\section{ALBL Experiments}
%As mentioned, the \texttt{ActiveLearningByLearning} (ALBL) meta-algorithm assists proper selection of strategies on the fly.
The following experiments%
\footnote{See \texttt{examples/albl\_plot.py} for details.}
demonstrate the usefulness of the ALBL meta-algorithm for algorithm selection on three datasets (\textit{heart}, \textit{australian}, \textit{diabetes}), downloaded from the dataset page of LIBSVM~\citep{libsvm}. ALBL is used to choose between Uncertainty Sampling, Random Sampling, QUIRE and HintSVM.
%, which are four concrete query strategies implemented in \textsc{libact}.
We take linear SVM from \textsc{scikit-learn} as the \texttt{Model}.
%% binary classification datasets from the libsvm dataset \citep{libsvm_dataset}
%% The active learning models considered were
%% Uncertainty Sampling, random, QUIRE, and HintSVM. The supervised
%% learning algorithm was SVM from \textsc{scikit-learn} with a linear kernel.
Figure \ref{fig:albl} demonstrates that the achieved error rates of ALBL
and the best strategy are very close to each other in every dataset, and verifies the usefulness
of ALBL for strategy selection.
%ALBL performed close to the best algorithm in each dataset.

\section{Package Specialties}
The source code of \textsc{libact} is hosted on Github (\url{https://github.com/ntucllab/libact}) with an issue tracker so that all users and developers can report problems when using this package. The package is
publicly available under a looser form of the BSD license, and
can be installed from the Python Package Index repository by \texttt{pip install libact}. The dependencies are listed in \texttt{requirements.txt} to facilitate installation, and
%The dependencies for \textsc{libact} are
%\textsc{numpy}~\citep{numpy2011van}, \textsc{SciPy}~\citep{scipy}, \textsc{scikit-learn}~\citep{scikit-learn},
%\textsc{Cython}~\citep{cython2010behnel}, \textsc{matplotlib}~\citep{matplotlib:Hunter:2007},
%and \textsc{Lapacke}~\citep{lapack}.
%The dependent packages are all free under BSD, Python Software Foundation~(PSF) and Apache licenses.
API references are written in \textsc{numpy}-style docstrings.
%and the whole documentation
%can be ).
%which are open source
%Python documentation generator and syntax highlighter under BSD license.
Documentation is automatically built with
 \textsc{Sphinx} and \textsc{Pygments}
 and
 hosted on \url{http://libact.readthedocs.org/}.
 %and
 %after pushing a new commit to \textsc{libact}'s Github repository.
We have also written unit tests and set up a continuous integration
testing environment on Travis CI (\url{https://travis-ci.org/ntucllab/libact}).
%Each time a new commit is pushed to the repository, Travis CI creates
%a Linux virtual environment, builds \textsc{libact} on it, and runs the unit tests.
This significantly reduces integration problems and facilitates more rapid project development.

One peer package on active learning is \textsc{jclal} \citep{reyes2016jclal}, which is a broad-covering Java library that also solves general active learning problems. Other current libraries either comes with a narrower coverage of active learning algorithms \citep{activer} or has not been deeply documented
\citep{santos2014comparison}. \textsc{jclal} enjoys the specialty of supporting stream-based active learning, along with user-friendly utilities in its library design. On the other hand, \textsc{libact} enjoys the specialty of supporting algorithm/parameter selection (ALBL) and cost-sensitive active learning, along with broader coverage on active learning algorithms for binary and multi-class classification.
%\textsc{jclal} is implemented in Java while
Also, \textsc{libact} can be easily integrated with the growing machine learning ecosystem in Python.

%Lapacke modified BSD
%Numpy , BSD
%SciPy BSD
%scikit-learn BSD
%matplotlib Python Software Foundation (PSF) license
%Cython, Apache License
%Sphinx BSD
%Pygments BSD

\section{Conclusion}
\textsc{libact} is a Python package designed to make active learning easy. It
contains interfaces to integrate with applications and other
packages, implementations of diverse active learning strategies to achieve
decent performance, automatic strategy selection routine via the ALBL
meta-algorithm to assist the users, and infrastructure in terms of issue
tracker, documentation and continuous integration to keep the package growing.

%% It designs interfaces for the necessary components of a given active learning
%% problem. With multiple implementations of different active learning algorithms and
%% the integration of the meta-algorithm ALBL, \textsc{libact} provides a tool for
%% particle uses. \textsc{libact} also has the infrastructure for version control,
%% unit tests, documentation, a issue tracker set up, which is convenient for
%% future development.

\section{Acknowledgements}
This package is based upon work supported by the Air Force Office of
Scientific Research, Asian Office of Aerospace Research and Development
(AOARD) under award number FA2386-15-1-4012, and by the Ministry of Science
and Technology of Taiwan under number MOST 103-2221-E-002-149-MY3.

\bibliographystyle{unsrt}
\bibliography{libact}

\end{document}